# Active Sensing as Bayes-Optimal Sequential Decision-Making

Sheeraz Ahmad
Computer Science and Engineering Dept.
University of California San Diego
La Jolla, CA 92093

Angela J. Yu
Cognitive Science Dept.
University of California San Diego
La Jolla, CA 92093

## Abstract

Sensory inference under conditions of uncertainty is a major problem in both machine learning and computational neuroscience. An important but poorly understood aspect of sensory processing is the role of *active* sensing. Here, we present a Bayes-optimal inference and control framework for active sensing, C-DAC (Context-Dependent Active Controller). Unlike previously proposed algorithms that optimize abstract statistical objectives such as information maximization (Infomax) [Butko and Movellan, 2010] or one-step look-ahead accuracy [Najemnik and Geisler, 2005], our active sensing model directly minimizes a combination of *behavioral costs*, such as temporal delay, response error, and sensor repositioning cost. We simulate these algorithms on a simple visual search task to illustrate scenarios in which context-sensitivity is particularly beneficial and optimization with respect to generic statistical objectives particularly inadequate. Motivated by the geometric properties of the C-DAC policy, we present both parametric and non-parametric approximations, which retain context-sensitivity while significantly reducing computational complexity. These approximations enable us to investigate a more complex search problem involving peripheral vision, and we notice that the performance advantage of C-DAC over generic statistical policies is even more evident in this scenario.

## 1 Introduction

In the realm of symbolic problem solving, computers are sometimes comparable, or even better than, typical human performance. In contrast, in sensory processing, especially under conditions of noise, uncertainty, or non-stationarity, human performance is often still the gold standard [Martin et al., 2001, Branson et al., 2011]. One important tool the brain has at its disposal is *active sensing*, a goal-directed, context-sensitive control strategy that prioritizes sensing resources toward the most rewarding or informative aspects of the environment [Yarbus, 1967]. Most theoretical models of sensory processing presume *passiveness*, considering only how to represent or compute with given inputs, and not how to actively intervene in the input collection process itself, especially with respect to behavioral goals or environmental constraints. Having a formal understanding of active sensing is not only important for advancing neuroscientific progress but also for engineering applications, such as developing context-sensitive, interactive artificial agents.

The most well-studied aspect of human active sensing is saccadic eye movements, and early work suggests that saccades are attracted to *salient* targets that differ from surround in one or more of feature dimensions such as orientation, motion, luminance, and color contrast [Koch and Ullman, 1985, Itti and Koch, 2000]. This passive explanation does not take into account the fact that the observations made while attending the task can affect the fixations decisions that follow. More recently, there has been a shift to relax this constraint of passiveness, and the notion of saliency has been reframed probabilistically in terms of maximizing the informational gain (Infomax) given the spatial and temporal context [Lee and Yu, 2000, Itti and Baldi, 2006, Butko and Movellan, 2010]. Separately, in another active formulation, it has been proposed that saccades are chosen to maximize the greedy, one-step look-ahead probability of finding the target (greedy MAP), conditioned on self knowledge about visual acuity map [Najemnik and Geisler, 2005].

While both the Infomax and Greedy MAP algorithms brought a new level of sophistication – representing sensory processing as iterative Bayesian inference, quantifying the knowledge gain of different saccade choices, and incorporating knowledge about sensory noise – they are still limited in several key respects: (1) they optimize abstract computational quantities that do not directly relate to behavioral goals (eg, speed and accuracy) or task constraints (eg, cost of switching from one location to another); (2) relatedly, it is unclear how to adapt these algorithms to varying task goals (eg, locating someone in a crowd versus catching a moving object); (3) there is no explicit representation of time in these algorithms, and thus no means of trading off fixation duration or number of fixations with search accuracy. In the rest of the paper, we refer to Infomax and Greedy MAP as "statistical policies", in the sense that they optimize generic statistical objectives insensitive to behavioral objectives or contextual constraints.

In contrast to the statistical policies, we propose a Bayes-optimal inference and control framework for active sensing, which we call C-DAC (Context-Dependent Active Controller). Specifically, we assume that the observer aims to optimize a context-sensitive objective function that takes into account behavioral costs such as temporal delay, response error, and the cost of switching from one sensing location to another. C-DAC uses this objective to choose when and where to collect sensory data, based on a continually updated statistically optimal (Bayesian) representation of the sequentially collected sensory data. This framework allows us to derive behaviorally optimal procedures for making decisions about (1) where to acquire sensory inputs, (2) when to move from one observation location to another, and (3) how to negotiate the exploration-exploitation tradeoff between collecting additional data and terminating the observation process. We also compare the performance of C-DAC and the statistical policies under different task parameters, and illustrate scenarios in which the latter perform particularly poorly. Finally, we present two approximate value iteration algorithms, based on a low-dimensional parametric and non-parametric approximation of the value function, which retain context-sensitivity while significantly reducing computational complexity.

In Sec. 2, we describe in detail the C-DAC model. In Sec. 3, we apply the model to a visual search task, simulating scenarios where C-DAC achieves a flexible trade-off between speed, accuracy and effort depending on the task demands, whereas the statistical policies fall short – this forms experimentally testable predictions for future investigations. We also present approximate value-iteration algorithms, and an extension of the search problem that incorporates peripheral vision. We conclude with a discussion of the implications of this work, relationship to previous work, as well as pointers to future work (Sec. 4).

## 2 The Model: C-DAC

We consider a scenario in which the observer must produce a response based on sequentially observed noisy sensory inputs (e.g., identifying target location in a search task or scene category in a classification task), with the ability to choose *where* and *how long* to collect the sensory inputs.

### 2.1 Sensory Processing: Bayesian Inference

We use a Bayesian generative model to capture the observer's knowledge about the statistical relationship among hidden causes or variables and how they give rise to noisy sensory inputs, as well as prior beliefs of hidden variables. We assume that they use exact Bayesian inference in the *recognition model* to maintain a statistically optimal representation of the hidden state of the world based on the noisy data stream.

Conditioned on the target location ($s$, hidden) and the sequence of fixation locations ($\boldsymbol{\lambda}_t := \{\lambda_1, \ldots, \lambda_t\}$, known), the agent sequentially observes iid inputs ($\mathbf{x}_t := \{x_1, \ldots, x_t\}$):

$$p(\mathbf{x}_t|s; \boldsymbol{\lambda}_t) = \prod_{i=1}^{t} p(x_i|s; \lambda_i) = \prod_{i=1}^{t} f_{s,\lambda_i}(x_i) \quad (1)$$

where $f_{s,\lambda}(x_t)$ is the likelihood function. These variables can be scalars or vectors, depending on the specific problem.

In the *recognition model*, repeated applications of Bayes' Rule can be used to compute the iterative posterior distribution over the $k$ possible target locations, or the *belief state*:

$$\mathbf{p}_t := (P(s=1|\mathbf{x}_t; \boldsymbol{\lambda}_t), \ldots, P(s=k|\mathbf{x}_t; \boldsymbol{\lambda}_t))$$
$$\mathbf{p}_t^i = P(s=i|\mathbf{x}_t; \boldsymbol{\lambda}_t) \propto p(x_t|s=i; \lambda_t)P(s=i|\mathbf{x}_{t-1}; \boldsymbol{\lambda}_{t-1})$$
$$= f_{s,\lambda_t}(x_t)\mathbf{p}_{t-1}^i \quad (2)$$

where $\mathbf{p}_0$ is the prior belief over target location.

### 2.2 Action Selection: Bayes Risk Minimization

The action selection component of active vision is a stochastic control problem where the agent chooses the sensing location and the number of data points collected, and we assume the agent can optimize this process dynamically based on ongoing data collection and size of sensory data, but the exact consequence of each action is not perfectly known ahead of time.

The goal is to find a good decision policy $\pi$, which maps the augmented belief state $(\mathbf{x}_t, \boldsymbol{\lambda}_t)$ into an action $a \in A$, where A consists of a set of termination actions, stopping and choosing a response, and a set of continuation actions, obtaining data point from a certain observation location. The policy $\pi$ produces for each observation sequence $(x_1, \ldots, x_t, \ldots)$, a stopping time $\tau$ (number of data points observed), a sequence of fixation choices $\boldsymbol{\lambda}_\tau := (\lambda_1, \ldots, \lambda_\tau)$, and an eventual target choice $\delta$.

In the Bayes risk minimization framework, the optimization problem is formulated in terms of minimizing an *expected* cost function, $L_\pi := \mathbb{E}[l(\tau, \boldsymbol{\lambda}_\tau, \delta)]_{\mathbf{x},s}$, averaged over stochasticity in the true target location $s$ and the data samples $\mathbf{x}$. We assume that the cost incurred on each trial takes into account *temporal delay*, *switch cost (cost associated with each switch in sensing location)*, and *response error*, respectively. In accordance with the typical Bayes risk formulation of the sequential decision problem, we assume the cost function to be a linear combination of the relevant factors:

$$l(\tau, \delta; \boldsymbol{\lambda}_\tau, s) = c\tau + c_s n_\tau + \mathbf{1}_{\{\delta \neq s\}} \quad (3)$$

where $n_\tau$ is the total number of switches ($n_\tau := \sum_{t=1}^{\tau-1} \mathbf{1}_{\{\lambda_{t+1} \neq \lambda_t\}}$), $c$ parameterizes the cost of temporal delay, $c_s$ the cost of a switch, and unit cost for response errors is assumed (as we can always divide $c$ and $c_s$ by the appropriate constant to make it 1). The expected cost is $L_\pi := c\mathbb{E}[\tau] + c_s\mathbb{E}[n_s] + P(\delta \neq s)$, where the expectation is taken over $\tau$, $\boldsymbol{\lambda}$, $\delta$, and $\mathbf{x}_\tau$.

Bellman's dynamic programming equation [Bellman, 1952] tells us that the problem is optimized if at each time point, the agent chooses the action associated with the lowest expected cost (the *Q-factor* for that action), given his current knowledge or *belief state*, $\mathbf{p}_t$. The Q-factors for the stopping actions are straight forward: $\bar{Q}_t^i(\mathbf{p}_t, \boldsymbol{\lambda}_t) := \mathbb{E}[l(t, i)|\mathbf{p}_t, \boldsymbol{\lambda}_t] = ct + c_s n_t + (1-\mathbf{p}_t^i)$. Obviously, the best stopping action $\delta$ is to minimize the probability of error. Thus, the stopping cost associated with the optimal stopping action ($i^* := \operatorname{argmax}_i \mathbf{p}_t^i$) is:

$$\begin{aligned}\bar{Q}_t^*(\mathbf{p}_t, \boldsymbol{\lambda}_t) &:= \mathbb{E}[l(t, i^*)|\mathbf{p}_t, \boldsymbol{\lambda}_t] \\ &= ct + c_s n_t + (1-\mathbf{p}_t^{i^*})\end{aligned} \quad (4)$$

The Q-factor associated with each continuation action $j$ (continue sensing in location $j$) is:

$$\begin{aligned}Q_t^j(\mathbf{p}_t = \mathbf{p}, \boldsymbol{\lambda}_t) &:= c(t+1) + c_s(n_t + \mathbf{1}_{\{j \neq \lambda_t\}}) + \\ &\min_{\tau', \delta, \boldsymbol{\lambda}_{\tau'}} \mathbb{E}[l(\tau', \delta)|\mathbf{p}_0 = \mathbf{p}, \lambda_1 = j]\end{aligned} \quad (5)$$

with the optimal continuation action being $Q_t^* := \min_j Q_t^j = Q_t^{j^*}$. The expected cost of continuing observing in location $j$ is equivalent to solving the original optimization problem with the prior belief set to the posterior after the previous $t$ time-steps, and the first observation location being $j$. Suppose we define the *value function* $V(\mathbf{p}, i)$ as the expected cost associated with the optimal policy, given prior belief $\mathbf{p}_0 = \mathbf{p}$ and initial observation location $\lambda_1 = i$:

$$V(\mathbf{p}, i) := \min_{\tau, \delta, \boldsymbol{\lambda}_\tau} \mathbb{E}[l(\tau, \delta)|\mathbf{p}_0 = \mathbf{p}, \lambda_1 = i] \ . \quad (6)$$

Then the value function satisfies the following recursive relation:

$$\begin{aligned}V(\mathbf{p}, k) &= \min(\bar{Q}_1^*(\mathbf{p}, k), Q_1^*(\mathbf{p}, k)) \\ &= \min\Big(\big(\min_i \bar{Q}_1^i(\mathbf{p}, k)\big), \\ &\quad \min_j \big(c + c_s \mathbf{1}_{\{j \neq k\}} + \mathbb{E}[V(\mathbf{p}', j)]\big)\Big)\end{aligned} \quad (7)$$

where $\mathbf{p}'$ is the belief state at next time-step, and the expectation is taken over the stochasticity in the next observation $x$. The optimal policy effectively divides the belief state space into a *stopping region* ($\bar{Q}^* \leq Q^*$) and a *continuation region* ($\bar{Q}^* > Q^*$), each of which further divided into subregions corresponding to alternative continuation and stopping actions. Note that the optimal decision policy is a *stationary* policy: the value function depends only on the belief state and observation location at the time the decision is to be taken, and not on time $t$ *per se*.

Bellman's dynamic programming principle implies a numerical algorithm for computing the optimal policy: guess an initial setting $V'(\mathbf{p}, k)$ of the value function (e.g., minimal stopping cost associated with each belief state $\mathbf{p}$ and observation location $k$), then iterate Eq. 7 until convergence, which yields the value function $V(\mathbf{p}, k) = V^\infty(\mathbf{p}, k)$.

## 3 Case Study: Visual Search

In this section, we apply the active sensing model to a simple, three location visual search task, where we can compute the exact optimal policy (up to discretization of the state space), and compare its performance with the statistical policies [Butko and Movellan, 2010, Najemnik and Geisler, 2005]. The target and distractors differ in terms of the likelihood of observations received, when looking at them.

### 3.1 C-DAC Policy

For simplicity, we assume that the observations are binary and Bernoulli distributed (iid conditioned on target and fixation locations):

$$p(x|s=i; \lambda_t = j) = \mathbf{1}_{\{i=j\}} \beta_1^x (1-\beta_1)^{1-x} + \mathbf{1}_{\{i \neq j\}} \beta_0^x (1-\beta_0)^{1-x}$$

The difficulty of the task is determined by the discriminability between target and distractor, or the difference between $\beta_1$ and $\beta_0$. For simplicity, we assume that the only stopping action available is to choose the current fixated location: $\hat{s}(\tau; \lambda_\tau = j) = j$. To reduce the parameter space, we also set $\beta_0 = 1 - \beta_1$, which is a reasonable assumption stating that the distractor and target stimuli only differ in one way (e.g. opposing direction of motion when using random dots stimulus with the coherence of dots kept the same). In the following, we first present a brief description of the greedy MAP and the infomax algorithms, before moving on to model comparisons.

### 3.2 Greedy MAP Policy

The *greedy MAP* algorithm [Najemnik and Geisler, 2005] suggests that agents should try to maximize the expected one-step look-ahead probability of finding the target. Thus, the reward function is:

$$R^g(\mathbf{p}_t, j) = \mathbb{E}_{x_{t+1}}[\max_i P(s = i | \mathbf{x}_t, x_{t+1}, \boldsymbol{\lambda}_t, \lambda_{t+1} = j)]$$
$$= \mathbb{E}_{x_{t+1}}[\max_i (\mathbf{p}_{t+1}^i) | x_{t+1}, \lambda_{t+1} = j]$$

To keep the notations consistent, we define the associated *Q-factor*, cost and policy as:

$$Q^g(\mathbf{p}_t, j) = -R^g(\mathbf{p}_t, j)$$
$$V^g(\mathbf{p}_t, j) = \min_j Q^g(\mathbf{p}_t, j)$$
$$\lambda_{t+1}^g = \operatorname{argmin}_j Q^g(\mathbf{p}_t, j)$$

### 3.3 Infomax Policy

The *infomax* algorithm [Butko and Movellan, 2010] tries to maximize the information gained from each fixation, by minimizing the expected cumulative future entropy. Similar to [Butko and Movellan, 2010], we can define the *Q-factors*, cost and the policy as:

$$Q^{im}(\mathbf{p}_t, j) = \sum_{t'=t+1}^{T} \mathbb{E}_{x_{t'}}[H(\mathbf{p}_{t'}) | x_{t'}, \lambda_{t+1} = j]$$
$$V^{im}(\mathbf{p}_t, j) = \min_j Q^{im}(\mathbf{p}_t, j)$$
$$\lambda_{t+1}^{im} = \operatorname{argmin}_j Q^{im}(\mathbf{p}_t, j)$$

where $H(\mathbf{p}) = -\sum_i \mathbf{p}^i \log \mathbf{p}^i$ is Shannon's entropy. Note that neither the original greedy MAP nor the infomax algorithm provide a principled answer as to when to stop searching and respond. They need to be augmented to stop once the maximum probability of any location containing the target exceeds a fixed threshold. We come back to the problem of how we set this threshold when we present comparison results.

### 3.4 Model Comparison

Before we discuss the performance of different models in terms of "behavioral" output, we first visually illustrate the decision policies (Fig. 1). The belief state $\mathbf{p}$ is represented by discretizing the two-dimensional belief state space $(\mathbf{p}^1, \mathbf{p}^2)$ with $m = 201$ bins in each dimension ($\mathbf{p}^3 = 1 - \mathbf{p}^1 - \mathbf{p}^2$). Although for C-DAC the policy also depends on the current fixation location, we only show it for fixating the first location; the other representations being rotationally symmetric. In Fig. 1, the parameters used for the C-DAC policy are $(c, c_s, \beta) = (0.1, 0, 0.9)$, and for the statistical policies, $(\beta, thresh) = (0.9, 0.8)$. Note that for this simple scenario with no switch cost, the infomax policy looks almost like the C-DAC policy – fixate the most likely location unless there is very strong evidence that the fixated location contains the target, in which case the observer should stop. The greedy MAP policy, on the other hand, looks completely different, and is in fact *ambiguous* in the sense that for a large set of belief states the policy does not give a unique next fixation location. We show one instance of this seemingly random policy, and note that there are regions where the policy suggests to look at either location 1 or 2 or 3 (corner regions speckled with green, orange and brown). Similarly, there are regions where the policy suggests to look at 1 or 2 (green+orange region). In fact, the performance of greedy MAP is so poor that we exclude it from the model comparisons below.

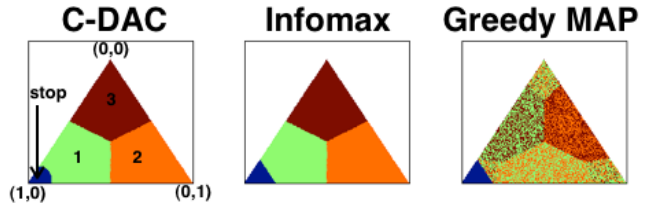

Figure 1: Decision policies – Infomax resembles C-DAC. Blue: stop. Green: fixate location 1. Orange: fixate location 2. Brown: fixate location 3. Environment $(c, c_s, \beta) = (0.1, 0, 0.9)$. Threshold for infomax and greedy MAP $= 0.8$

Fig. 2 shows the effects of how the C-DAC policy changes when different parameters of the task are changed. As seen in the figure, the stopping region expands if the cost of time increases (high $c$), intuitively this makes sense – if each time step is costlier then the observer should stop at a lower level of confidence, at the expense of higher error rate. Similarly, for the case when $\beta$ is smaller (high noise), stopping with a lower level of confidence makes sense – the value of each additional observation depends on how noisy the data is, the noisier the less worthwhile to continue observing, thus leading to a lower stopping criterion. Lastly, and arguably the most interesting case, is when there is an additional switch cost (added $c_s$); this deters the algorithm from switching even when the belief in a given location has reduced below $1/3$. In fact, this is the scenario where optimizing for behavioral objectives turns out to be truly beneficial, and although infomax can approximate the C-DAC policy when the switch cost is 0, it cannot do so when switch cost comes in to play.

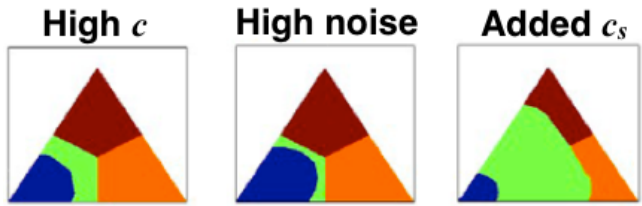

Figure 2: C-DAC policy for different environments $(c, c_s, \beta)$ – high $c$ $(0.2, 0, 0.9)$, high noise $(0.1, 0, 0.7)$, and added $c_s$ $(0.1, 0.1, 0.9)$.

Next, we look at how these intuitions from the policy plots translate to output measures in terms of accuracy, response delay, and number of fixations. In order to set the stopping threshold for the infomax policy in the most generous/optimistic setting, we first run the C-DAC policy, and then set the threshold for infomax so that it matches the accuracy of C-DAC [1], while we compare the other output measures. We choose two scenarios: (1) no switch cost, (2) with switch cost. For all simulations, the algorithm starts with uniform prior ($\mathbf{p} = (1/3, 1/3, 1/3)$) and initial fixation location 1, while the true target location is uniformly distributed. Fig. 3 shows the accuracy, number of time steps and number of switches for both scenarios. Confirming the intuition from the policy plots, the performance of infomax and C-DAC are comparable for $c_s = 0$. However, when a switch cost is added, $c_s = 0.2$, we see that although the accuracy is comparable by design, there is small improvement in search time of C-DAC, and a notable advantage in the number of switches. The behavior of the infomax policy does not adapt to the change in the behavioral cost function, thus incurring an overall higher cost. Algorithms like infomax that maximize abstract statistical objectives lack the inherent flexibility to adapt to changing behavioral goals or environmental constraints. Even for this simple visual search example, Infomax does not have a principled way of setting the stopping threshold, and we gave it the best-scenario outcome by adopting the stopping policy generated by C-DAC in different contexts.

---

[1] Since a binary search is required to set this matching threshold, and the accuracy is sensitive w.r.t. this threshold, we settle on an approximate accuracy match for infomax that is comparable or lower than C-DAC.

### 3.5 Approximate Control

Our model is formally a variant of POMDP (Partially Observable Markov Decision Process), or, more specifically, a Mixed Observability Markov Decision Process (MOMDP) [Ong et al., 2010, Araya-López et al., 2010], which differs from ordinary POMDP in that part of the state space is partly hidden (target location in our case) and partly observable (current fixation location in our case). In general, POMDPs are hard to solve since the decision made at each time step depends on all the past actions and observations, thus imposing enormous memory requirements. This is known as the curse of history, and is the first major hurdle towards any practical solution. An elegant way to alleviate this is to use belief states which serve as a sufficient statistic for the process history, thus requiring to maintain just a single distribution instead of the entire history. Converting a POMDP to a belief-state MDP is in fact a prevalent technique and the one we employ. However, this leads to another computational hurdle, known as the curse of dimensionality, since now we have a MDP with a continuous state-space, making tabular representation of value function infeasible. One way to work around the problem is to discretize the belief state space into a grid, where instead of finding the value function at all the points in the belief state simplex, we only do so for a finite number of grid points. The grid approximation, that we also use, has appealing performance guarantees which improve as the density of the grid is increased [Lovejoy, 1991]. To evaluate the value function at the points not in this set, we use some sort of interpolation technique (value at the nearest grid point, weighted average value at $k$-nearest grid point, etc.). However, although grid approximation may work for small state spaces, it does not scale well to larger, practical problems. For example, when used for the active sensing problem with $k$ sensing locations, a uniform grid of size $n$ has $O(kn^{k-1})$ complexity.

Although there is a rich body of literature on approximate solutions of POMDP (e.g. [Powell, 2007, Lagoudakis and Parr, 2003, Kaplow, 2010]) tackling both general as well as application-specific approximations, most are inappropriate for dealing with the MOMDP problem such as the one encountered here. Furthermore, most of the POMDP approximation algorithms focus on discounted rewards and/or finite-horizon problems. Our formulation does not fall into these categories and thus require novel approximation schemes. We note that the Q-factors and the resulting value function are smooth and concave, making them amenable to low dimensional approximations. At each step, we find a low dimensional representation of the value function, and use that for the update step of the value iteration algorithm. Specifically, instead of re-computing the value function at each grid point, here we generate a large number of samples uniformly on the belief state space, compute a new estimate of the value function at those locations, and then extrapolate the value function to everywhere by improving its parametric fit.

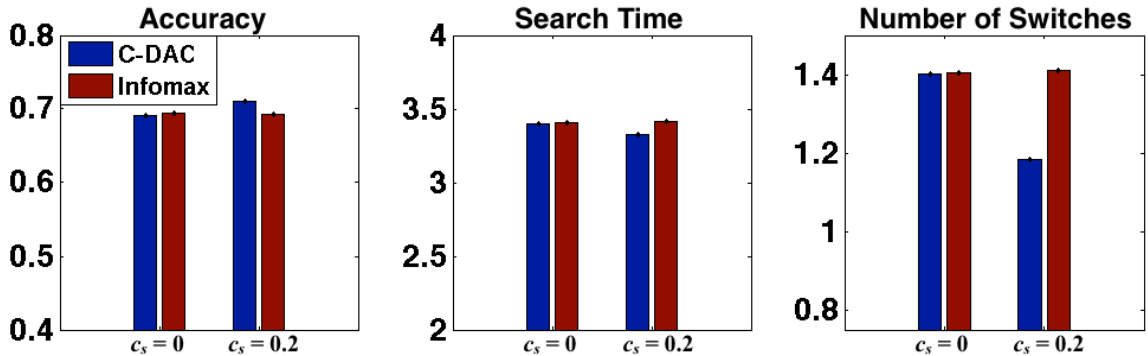

Figure 3: Comparison between C-DAC and Infomax for two environments $(c, c_s, \beta) = (0.1, 0, 0.8)$ and $(0.1, 0.2, 0.8)$. C-DAC has superior performance when $c_s > 0$.

The first low-dimensional approximation we consider is the Radial Basis Functions (RBF) representation:

1. Generate M RBFs, centered at $\{\mu_i\}_{i=1}^{M}$, with fixed $\sigma$: $\phi(\mathbf{p}) = \frac{1}{\sigma(2\pi)^{k/2}} e^{\frac{||\mathbf{p}-\mu_i||^2}{2\sigma^2}}$
2. Generate $m$ random points from belief space, $\mathbf{p}$.
3. Initialize $\{V(\mathbf{p}_i)\}_{i=1}^{m}$ with the stopping costs.
4. Find minimum-norm $\mathbf{w}$ from: $V(\mathbf{p}) = \Phi(\mathbf{p})\mathbf{w}$.
5. Generate new $m$ random belief state points ($\mathbf{p}'$).
6. Evaluate required $V$ values using current $\mathbf{w}$.
7. Update $V(\mathbf{p}')$ using value iteration.
8. Find a new $\mathbf{w}$ from $V(\mathbf{p}') = \Phi(\mathbf{p}')\mathbf{w}$.
9. Repeat steps 5 through 8, until $w$ converges.

While we adopt a Gaussian kernel function, other constructs are possible and have been implemented in our problem without significant performance deviation (not shown), e.g. multiquadratic ($\phi(\mathbf{p}) = \sqrt{1 + \epsilon ||\mathbf{p} - \mu_i||^2}$), inverse-quadratic($\phi(\mathbf{p}) = (1 + \epsilon ||\mathbf{p} - \mu_i||^2)^{-1}$), thin plate spine ($\phi(\mathbf{p}) = ||\mathbf{p} - \mu_i||^2 \ln ||\mathbf{p} - \mu_i||$), etc. [Buhmann, 2003].

The RBF approximation requires setting several parameters (number, mean, and variance of bases), which can be impractical for large problems, when there is little or no information available about the properties of the true value function. We thus also implement a nonparametric variation of the algorithm, whereby we use Gaussian Process Regression (GPR) [Williams and Rasmussen, 1996] to estimate the value function (step 4, 6 and 8). In addition, we also implement GPR with hyperparameter learning (Automatic Relevance Determination, ARD), thus obviating the need to pre-set model parameters.

The approximations lead to considerable computational savings. The complexity of the RBF approximation is $O(k(mM + M^3))$, for $k$ sensing locations, $m$ random points chosen at each step, and $M$ bases. For the GPR approximation, the complexity is $O(kN^3)$, where $N$ is the number of points used for regression. In practice, all the approximation algorithms we consider converge rapidly (under 10 iterations), though we do not have a proof that this holds for a general case.

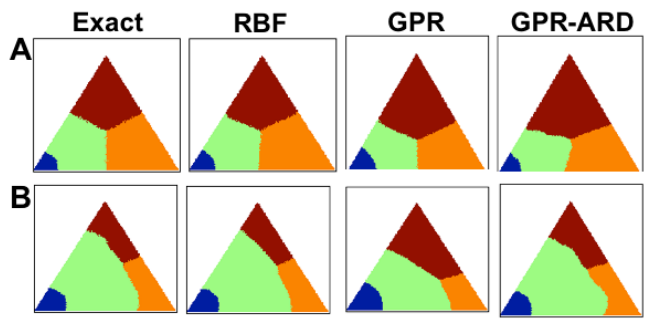

Figure 4: Exact vs. approximate policies shown over $n = 201$ bins. (A) Environment $(c, c_s, \beta) = (0.1, 0, 0.9)$. (B) $(c, c_s, \beta) = (0.1, 0.1, 0.9)$.

In the simulations, the RBF approximate policy uses $m = 1000$ random point for each iteration, and $M = 49$ bases, uniformly placed in the belief simplex, with a unit variance. The GPR approximate policy uses a unit length scale, unit signal strength and a noise-strength of 0.1, with $N = 200$ random points used for regression. Fig. 4A shows the exact policy vs. the learned approximate policies for different approximations when the switch cost is 0, $(c, c_s, \beta) = (0.1, 0, 0.9)$. We notice that with handcrafted bases, RBF is a good approximation of the exact policy, whereas relaxing the parametric form in GPR and subsequently learning the hyperparameters in GPR with ARD, leads to a slightly poorer but more robust non-parametric approximation. Similar observations can be made in Fig. 4B, for the environment with added switch cost, $(c, c_s, \beta) = (0.1, 0.1, 0.9)$. All the results are shown over a 201x201 grid. These faster yet robust approximations motivated us to apply our model to more complex problems. We investigate one such problem of visual search with peripheral vision next, and show how our model is fundamentally different from existing formulations such as infomax, even when the cost of effort is not considered.

### 3.6 Visual Search with Peripheral Vision

In the very simple three-location visual search problem we considered above, we did not incorporate the possibility of peripheral vision, or the more general possibility that a sensor positioned in a particular location can have distance-dependent, degraded information about nearby locations as well. We therefore consider a simple example with peripheral vision (see Fig. 5B), whereby the observer can saccade to intermediate locations that give reduced information about either two (sensing locations on the edges of the triangle) or three (sensing location in the center) stimuli. This is motivated by experimental observations that humans not only fixate most probable target locations but sometimes also center-of-gravity locations that are intermediate among two or more target locations [Findley, 1982, Zelinsky et al., 1997].

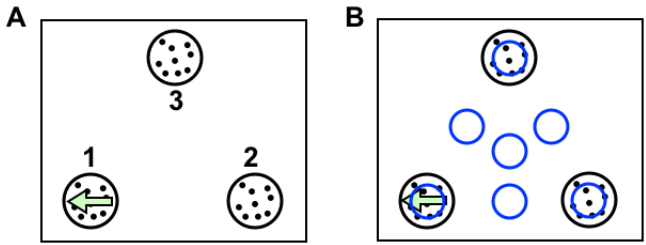

Figure 5: Schematics of visual search task. The general task is to find the target (left-moving dots) amongst distractors (right-moving dots). Not drawn to scale. (A) Task 1: agent fixates one of the target patches at any given time. (B) Task 2: agent fixates one of the blue circle regions at any given time

Formally, we need an *acuity map*, the notion that it is possible to gain information about stimuli peripheral to the fixation center (fovea), such that the quality of that information decays at greater spatial distance away from the fovea. For example, the task of Fig. 5B would require a continuation action space of 7 elements, $L = \{l_1, l_2, l_3, l_{12}, l_{23}, l_{13}, l_{123}\}$, where the first three actions correspond to fixating one of the three target locations, the next three to fixating midway between two target locations, and the last to fixating the center of all three. We parameterize the quality of peripheral vision by augmenting the observations to be three-dimensional, $(x^1, x^2, x^3)$, corresponding to the three simultaneously viewed locations. We assume that each $x_i$ is generated by a Bernoulli distribution favoring 1 if it is the target, and 0 if it is not, and its magnitude (absolute difference from 0.5) is greatest when observer directly fixates the stimulus, and smallest when the observer directly fixates one of the other stimuli. We use 4 parameters to characterize the observations $(1 > \beta_1 > \beta_2 > \beta_3 > \beta_4 >= 0.5)$. So, when the agent is fixating one of the potential target locations ($l_1$, $l_2$ or $l_3$), it gets an observation from the fixated location (parameter $\beta_1$ or $1 - \beta_1$ depending on whether it is the target or a distractor), and observations from the non-fixated locations (parameter $\beta_4$ or $1 - \beta_4$ depending on whether they are a target or a distractor). Similarly, for the midway locations ($l_{12}$, $l_{23}$ or $l_{13}$), the observations are received for the closest locations (parameter $\beta_2$ or $1 - \beta_2$ depending on whether they are a target or a distractor), and from the farther off location (parameter $\beta_4$ or $1 - \beta_4$ depending on whether it is the target or a distractor). Lastly, for the center location ($l_{123}$), the observations are made for all three locations (parameter $\beta_3$ or $1-\beta_3$ depending on whether they are a target or a distractor). Furthermore, since the agent can now look at locations that cannot be target, we relax the assumption that the agent must look at a particular location before choosing it, allowing the agent to stop at any location and declare the target.

### 3.7 Model Comparison

We first present the policies, and, similar to our discussion of simple visual search task, we only show the C-DAC policy looking at the first location ($l_1$) (the other fixation-dependent policies are rotationally symmetric). It is evident from Fig. 6 that now the C-DAC policy differs from the infomax policy even when no switch cost is considered, thus pointing to a more fundamental difference between the two. Note that for the parameters used here, C-DAC never chooses to look at the center $l_{123}$, but it does so for other parameter settings (not shown). Infomax, however, never even looks at the actual potential locations, favoring only midway locations before declaring the target location.

For performance comparison in terms of behavioral output, we again investigate two scenarios: (1) no switch cost, (2) with switch cost. The threshold for infomax is set so that the accuracies are matched to facilitate fair comparison. For all simulations, the al-gorithm starts with uniform prior ($\mathbf{p} = (1/3, 1/3, 1/3)$) and initial fixation at the center (location $l_{123}$), while the true target location is uniformly distributed. Fig. 7 shows the accuracy, number of time steps, and number of switches for both scenarios. Now we notice that C-DAC outperforms infomax even when switch cost is not considered, in contrast to the simple task without peripheral vision (Fig. 3). Note however that C-DAC makes more switches for $c_s = 0$, which makes sense since switches have no cost, and search time can potentially be reduced by allowing more switches. However, when we add a switch cost ($c_s = 0.005$), C-DAC significantly reduces number of switches, whereas infomax lacks this adaptability to a changed environment.

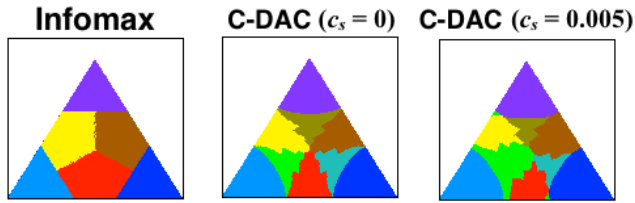

Figure 6: Decision policies. Azure: stop and choose location $l_1$. Blue: stop and choose $l_2$. Indigo: stop and choose $l_3$. Green: fixate $l_1$. Sea-green: fixate $l_2$. Olive: fixate $l_3$. Red: fixate $l_{12}$. Brown: fixate $l_{23}$. Yellow: fixate $l_{13}$. Environment $(c, \beta 1, \beta_2, \beta_3, \beta_4) = (0.05, 0.62, 0.6, 0.55, 0.5)$. Threshold for infomax $= 0.6$

## 4 Discussion

In this paper, we proposed a POMDP plus Bayes risk-minimization framework for active sensing, which optimizes behaviorally relevant objectives in expectation, such as speed, accuracy, and switching efficiency. We compared this C-DAC policy to the previously proposed *infomax* and *greedy MAP* policies. We found that greedy MAP performs very poorly, and although Infomax can approximate the optimal policy for some simple environments, it lacks intrinsic context sensitivity or flexibility. Specifically, for different environments, there is no principled way to set a decision threshold for either greedy MAP or Infomax, leading to higher costs, longer fixation durations, and larger number of switches in problem settings when those costs are significant. This performance difference and the advantage of the added flexibility provided by C-DAC becomes even more profound when we consider a more general visual search problem with peripheral vision. The family of approximations that we present opens up the avenue for application of our model to complex, real world problems.

There have been several other related active sensing algorithms that differ from C-DAC in their state representation, inference, control and/or approximation scheme. We briefly summarize some of these here. In [Darrell and Pentland, 1996], the problem of active gesture recognition is studied, by using historic state representation and nearest neighbor Q-function approximation. Sensing strategies for robots in RoboCup competition is studied in [Kwok and Fox, 2004], which uses states augmented with associated uncertainty and model-free Least Square Policy Iteration (LSPI) approximation [Lagoudakis and Parr, 2003]. Context dependent goals are considered in [Ji et al., 2007] and [Naghshvar and Javidi, 2010]. The former concentrates on multi-sensor multi-aspect sensing using Point Based Value Iteration (PBVI) approximation [Pineau et al., 2006]. The latter aims to provide conditions for reduction of an active sequential hypothesis testing problem to passive hypothesis testing. A Reinforcement Learning paradigm where reward is not dependent on information gain but on how close a saccade brings the target to the optical axis has also been proposed [Minut and Mahadevan, 2001]. Other control strategies like random search, sequential sweeping search, "Drosophila-inspired" search [Chung and Burdick, April 2007] and hierarchical POMDPs for visual action planning [Sridharan et al., 2010] have also been proposed. We choose infomax to compare our C-DAC policy against because, as a human-vision inspired model, it not only explains human fixation behavior on a variety of tasks, but also has cutting edge computer vision applications (e.g. the digital eye [Butko and Movellan, 2010]).

A related problem domain, not typically studied as POMDP or MDP, is Multi-Armed Bandits (MAB) [Gittins, 1979]. The classical example of a MAB problem concerns with pulling levers (or playing arms) in a set of slot machines. The person gambling is unaware of the states and reward distribution of the levers, and has to figure out which lever to pull next in order to maximize the cumulative reward. Noting a correspondence between the ideas of pulling arms and fixating location, and between rewards and observations, the MAB framework seems to describe the active sensing problem. Concretely, given the locations fixated (arms played) so far, and the observations (rewards) received, how to choose which location to fixate (which arm to play) next. However, there are certain characteristics of the active sensing problem that make it difficult to study in a MAB framework as yet. Firstly, the problem is an instance of restless bandits [Whittle, 1988], where the state of an arm can change even when it is not played. In active sensing, the belief about a location being the target does change even when it is not fixated. Whittles index is a simple rule that assigns a value to each arm in a restless setting, and the arm with the highest value is then played. The rule is asymptotically optimal only for a sub-class of problems (e.g. [Washburn and Schneider, 2008] and [Liu and Zhao, 2010]), but not optimal in general. Secondly, the states of the arms in the active sensing task are correlated (the elements of the belief-state have to add up to 1). There is some work on correlated arms for specific structure of correlation, like clustered arms [Pandey et al., 2007] and Gaussian process bandits [Dorard et al., 2009], but so far there is no general strategy for handling this scenario.

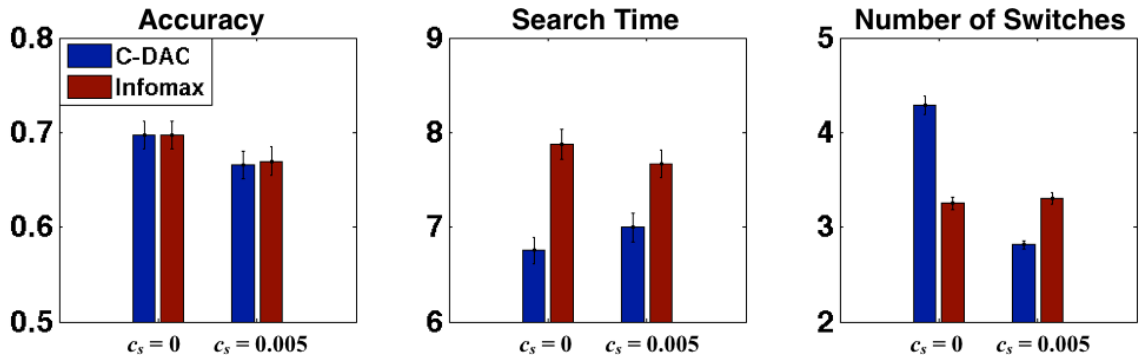

Figure 7: Comparison between C-DAC and Infomax on Task 2 for two environments $(c, \beta 1, \beta_2, \beta_3, \beta_4) = (0.05, 0.62, 0.6, 0.55, 0.5)$, $c_s = 0$ and $c_s = 0.005$. C-DAC adjusts the time steps and number of switches depending on the environment, taking a little longer but reducing number of switches when effort has cost.

Active learning is another related approach, with hypothesis testing as a sub-problem that is related to the problem of active sensing. The setting involves an unknown true hypothesis, and an agent that can perform queries providing information about the underlying hypothesis. The task is then to determine which query to perform next to optimally reduce the number of plausible hypothesis (version space). In active sensing however, although the belief about a hypothesis (target location) can become arbitrarily low, the number of plausible hypothesis does not reduce. This problem is investigated in [Golovin et al., 2010], and a near-optimal greedy solution is proposed along with performance guarantees. Besides the sub-optimality of the approach, the same test cannot be performed more than once (whereas in active sensing, one location can be fixated more than once). The lack of this provision stems from the fact that the noisy observations considered are actually deterministic with respect to a hidden noise parameter. Thus, as of yet it is hard to cast the active sensing problem in this framework.

We thus conclude that although there is a rich body of literature on related problems, as can be seen from the few examples we presented, our formulation is novel (to our best knowledge) in its goals and principled approach to the problem of active sensing. In general, the framework proposed here has the potential for not only applications in visual search, but a host of other problems, ranging from active scene categorization to active foraging. The decision policies it generates are adaptive to the environment and sensitive to contextual factors. This flexibility and robustness to different environments makes the framework an appealing choice for a variety of active sensing applications.